\documentclass[a4paper, 10pt, conference]{ieeeconf}      % Use this line for a4 paper

\IEEEoverridecommandlockouts                              % This command is only needed if 
\usepackage{filecontents}

\usepackage{xcolor}
\usepackage{float}
\usepackage{hyperref}
\usepackage{blindtext}
\usepackage{pdfpages}
\usepackage{pgfplots}
\usepackage{tikz}
\usepackage{tikzscale}
\usepackage{graphicx}
\usepackage{pgf}
\usepackage{color}
\usepackage{etoolbox}
\usepackage{graphicx}
\usepackage{subcaption}
\usepackage{caption}
\usepackage{scalefnt}
\usepackage{standalone}
\usepackage{array}
\usepackage{booktabs}
\usetikzlibrary{positioning}
\pgfdeclarelayer{bg}    % declare background layer
\pgfsetlayers{bg,main}
\usetikzlibrary{positioning,shapes,arrows,shapes,calc,trees,decorations,shadows}
\usepackage[acronym,automake]{glossaries-extra}
\usepackage{amsmath}
\usepackage{amsfonts} % for the \checkmark command 
\usepackage{siunitx} % SI units
\usepackage[font=small]{caption} %TODO: check if this applies to ITSC
\def\equationautorefname#1#2\null{({#2\null})} % Define equation usage as (n)

\renewcommand{\figureautorefname}[1]{Fig.\,}

\setabbreviationstyle[acronym]{long-short}

\loadglsentries{example-glossaries-brief}
\loadglsentries{example-glossaries-acronym}

\newacronym{CNN}{CNN}{convolutional neural network}
\newacronym[longplural={long short-term memories}]{LSTM}{LSTM}{long short-term memory}
\newacronym{GMM}{GMM}{Gaussian mixture model}
\newacronym{HMM}{HMM}{hidden Markov model}
\newacronym{ReLu}{ReLu}{rectified linear unit}
\newacronym{RCNN}{RCNN}{region-based convolutional neural network}
\newacronym{SVM}{SVM}{support vector machine}
\newacronym{FC}{FC}{fully-connected} \newacronym{R-CNN}{R-CNN}{regions with CNN features} \newacronym{CDC}{CDC}{convolutional-de-convolutional}
\newacronym{UTM}{UTM}{Universal Transverse Mercator}
\newacronym{GIS}{GIS}{geographic information system}
\newacronym{XML}{XML}{extensible markup language}
\newacronym{MH}{MH}{modified Hausdorff}
\usepackage{blindtext}
\title{\LARGE \bf
  openDD: A Large-Scale Roundabout Drone Dataset
}

\author{Antonia Breuer$^{1}$, Jan-Aike Term\"ohlen$^{2}$, Silviu Homoceanu$^{1}$, Tim Fingscheidt$^{2}$%
\thanks{$^{1}$Antonia Breuer and Silviu Homoceanu are with Volks\-wagen, Berliner Ring 2, 38440 Wolfsburg, Germany.\,\,\texttt{\small \{antonia.breuer, silviu.homoceanu\}@volkswagen.de}}%
\thanks{$^{2}$Jan-Aike Term\"ohlen and Tim Fingscheidt are with the Institute for Communications Technology, Technische Universit\"at Braunschweig, Schleinitzstr. 22, 38106 Braunschweig, Germany.
{\tt\small j.termoehlen, t.fingscheidt@tu-bs.de}}%
}

\makeglossaries
\begin{document}
\maketitle
\thispagestyle{empty}
\pagestyle{empty}

%%%%%%%%%%%%%%%%%%%%%%%%%%%%%%%%%%%%%%%%%%%%%%%%%%%%%%%%%%%%%%%%%%%%%%%%%%%%%%%%
\begin{abstract}
  Analyzing and predicting the traffic scene around the ego vehicle has been one of the key challenges in autonomous driving.
  Datasets including the trajectories of all road users present in a scene, as well as the underlying road topology are invaluable to analyze the behavior of the various traffic participants.
  The interaction between the traffic participants is especially high in intersection types that are not regulated by traffic lights, the most common one being the roundabout.
  We introduce the \texttt{openDD} dataset, including 84,774 accurately-tracked trajectories and HD map data of seven different roundabouts.
  The \texttt{openDD} dataset is annotated using images taken by a drone in 501 separate flights, totalling in over 62 hours of trajectory data.
  As of today the \texttt{openDD} is by far the largest publicly available trajectory dataset recorded from a drone perspective, while comparable datasets span 17 hours at most.
  The data is available, for both commercial and non-commercial use, at: \mbox{\url{http://www.l3pilot.eu/openDD}}.
\end{abstract}

%%%%%%%%%%%%%%%%%%%%%%%%%%%%%%%%%%%%%%%%%%%%%%%%%%%%%%%%%%%%%%%%%%%%%%%%%%%%%%%%
\section{INTRODUCTION} \glsresetall
In recent years autonomous driving has become one of the major applications for numerous fields of research.
A main challenge faced in autonomous driving is the prediction of the traffic scene surrounding the ego vehicle.
Predicting the surrounding traffic scene is particularly difficult in urban and rural scenarios due to the high inter-dependency between the involved road users.
One way to face this challenge is to use increasing amounts of data, causing a surge of popularity of data-driven approaches that rely on large-scale datasets in recent years~\cite{ridel_literature_2018, cui_multimodal_2018, breuer_analysis_2019, cui_deep_2019}.
Other applications of trajectory datasets include the modeling and analysis of driving behavior~\cite{driggs-campbell_integrating_2017} and the analysis of safety of the autonomous driving function~\cite{noauthor_enable-s3_nodate}.
\begin{figure}[t]
  \includegraphics[width=\linewidth]{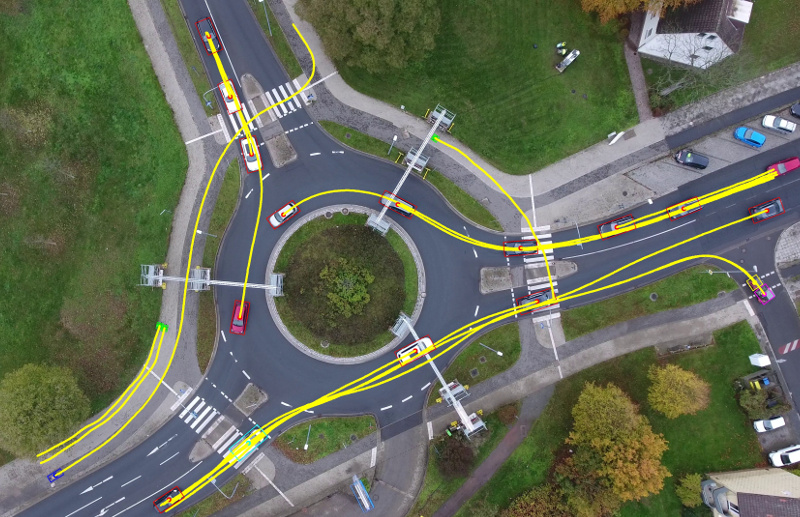}
  \caption{
    An exemplary visualization of a given traffic scene included in the dataset.
    The past trajectories of all present objects relative to the given time instant $n$ are drawn in yellow.
    The bounding box of object $j$ is depicted in the color corresponding to its class $c_n^{(j)}$.
   }
   \label{fig:visualizedTrajectories}
\end{figure}

Datasets recorded from a ground view instead of a bird's eye perspective are limited by occlusions and the restricted field of view of the recording device at the ground.
Labeling the trajectories with the help of image data captured by an aerial drone ensures a complete overview of the traffic situation and enables algorithms that use the dataset to take all present road users into account.
% Especially in urban scenarios a road user's behavior is heavily influenced by the behavior of others.
The interaction between different road users is particularly high in intersections that are not regulated by traffic lights, the most common being the roundabout.
In the \texttt{openDD} dataset presented in this work, seven roundabouts with different topologies are covered, one of them shown in \autoref{fig:visualizedTrajectories}.
The dataset includes trajectories of all recorded road users, shapefiles and an \gls{XML} file, describing the road topology of the underlying intersections.
One reference image taken by the drone is provided per intersection.
An exemplary visualization of the data included in the dataset can be seen in \autoref{fig:visualizedTrajectories}.
As shown there, all dynamic traffic participants are accurately tracked in the relevant area surrounding the roundabout, and also vehicles hard to see for the human eye, such as the grey car on top of the picture are accurately detected.
The introduced \texttt{openDD} dataset spans more than $62$ hours in total, covers $84,774$ trajectories, and can be accessed on the following website: \mbox{\url{http://www.l3pilot.eu/openDD}}.

\section{RELATED WORK} \label{sec:relatedWork}
\begin{table*}[h!]
  \caption{Overview of trajectory datasets recorded by a drone published in recent years.} \label{tab:relatedWork_datasets}
  \begin{center}
    \begin{tabular}{c|r|l|c|r|c|l|l}
      \multicolumn{1}{c|}{Dataset Name} & \multicolumn{1}{c|}{Length} & Situations & \multicolumn{1}{c|}{\# locations} &\multicolumn{1}{c|}{\# trajectories} & \multicolumn{1}{c|}{Map} & Classes & License \\
      \hline
      \hline
      \texttt{Stanford} & \SI{9,00}{\hour} & campus & \SI{}{8} & 10,240 & none & pedestrians, bicycles, cars, & non-commercial \\
       \texttt{\quad Drone}~\cite{robicquet_learning_2016}              & &        &   &        &      & skateboards, carts, buses & \\
      \hline
    \texttt{highD}~\cite{krajewski_highd_2018} & \SI{16,50}{\hour} & highway & 6 & 110,000 & none & cars, trucks & non-commercial \\
      \hline
      \texttt{CITR}~\cite{yang_top-view_2019} & \SI{0,21}{\hour} & parking lot & 1 & 340 & none & pedestrians & non-commercial \\
      \hline
      \texttt{DUT}~\cite{yang_top-view_2019} & \SI{0,16}{\hour} & urban intersections, & 2 & 1,793 & none & pedestrians & non-commercial \\
      & & shared space & & & & & \\
      \hline
    \texttt{inD}~\cite{bock_ind_2019} & \SI{10,00}{\hour} & urban intersections & 4 & 11,500 &  none & pedestrians, bicycles, cars, & non-commerical \\
                                 &     &                     &   &       & & trucks, buses               &\\
      \hline
      \texttt{INTERACTION} & \SI{16,50}{\hour} & urban intersections, & 11 & 40,054 & \texttt{lanelet2} & cars, pedestrians  & non-commerical \\
      \cite{wei_interaction_2019} & & highway & & & \cite{poggenhans_lanelet2:_2018} & & \\
      \hline
      \texttt{openDD} & \SI{62,70}{\hour}& roundabouts & 7 & 84,774 & shapefiles & cars, vans, trucks, buses, & non-commerical \\
             &     &                     &   &     &                                           & pedestrians, trailers,& \& commercial \\
             &     &                     &   &     &                                           & motorcycles, bicyclists & \\
    \end{tabular}
  \end{center} \end{table*}

\autoref{tab:relatedWork_datasets} gives an overview of trajectory datasets recorded from a drone perspective and their characteristics.

The \texttt{Stanford drone} dataset~\cite{robicquet_learning_2016} was the first publicly available trajectory dataset recorded from a drone's perspective and is tailored to the analysis of pedestrian trajectories.
It consists of $9$ hours of data over eight unique locations on the campus and has a high percentage rate of labeled pedestrians and cyclists, while only about $7\%$ of the labeled targets are cars.
The \texttt{DUT} and \texttt{CITR} datasets~\cite{yang_top-view_2019} are especially designed for the analysis of the behavior of pedestrians when interacting with vehicles and span less than half an hour in total.
One of the first large-scale trajectory datasets based on the footage of an aerial drone is the \texttt{highD} dataset~\cite{krajewski_highd_2018}.
It includes trajectory data from $110,000$ cars on German highways and spans $5,600$ lane changes over $16.5$ hours of data.
During the creation of the presented \texttt{openDD} dataset, descriptions of the \texttt{INTERACTION}~\cite{wei_interaction_2019} and the \texttt{inD}~\cite{bock_ind_2019} datasets have been published.
The \texttt{INTERACTION} dataset spans about $16.5$ hours and covers data from $11$ intersections, including $5$ roundabouts, $3$ unsignalized intersections, $2$ merging and lane change situations, and $1$ signalized intersection.
The \texttt{InD} dataset~\cite{bock_ind_2019} distinguishes pedestrians, bicycles, cars, trucks, and buses and includes 10 hours of data recorded by a drone.
At the time of writing of this publication, the \texttt{InD} dataset has not been released yet, thus no further description than the one stated in the publication can be given.
% TODO: control Stanford drone dataset length
% TODO: control DUT dataset length

\section{DATASET} \label{sec:maneuverDetection}
\begin{figure}[h]
% \vspace{0.5em}

	\begin{subfigure}{0.32\linewidth}
	   \centering
	   \includegraphics[width=\linewidth]{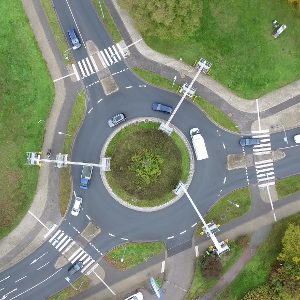}
       \caption*{$\text{rdb}_1$}
	\end{subfigure}
	\begin{subfigure}{0.32\linewidth}
	   \centering
	   \includegraphics[width=\linewidth]{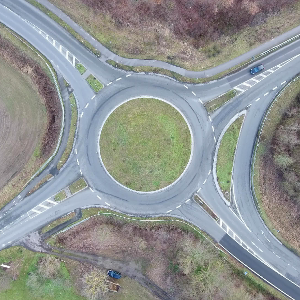}
       \caption*{$\text{rdb}_2$}
	\end{subfigure}
	\begin{subfigure}{0.32\linewidth}
	   \includegraphics[width=\linewidth]{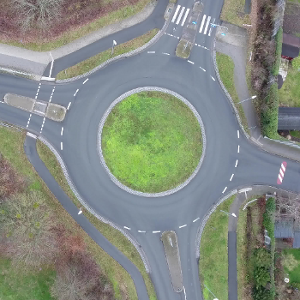}
       \caption*{$\text{rdb}_3$}
	\end{subfigure}
	\begin{subfigure}{0.32\linewidth}
	   \includegraphics[width=\linewidth]{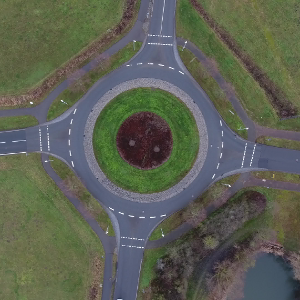}
       \caption*{$\text{rdb}_4$}
	\end{subfigure}
	\begin{subfigure}{0.32\linewidth}
	   \includegraphics[width=\linewidth]{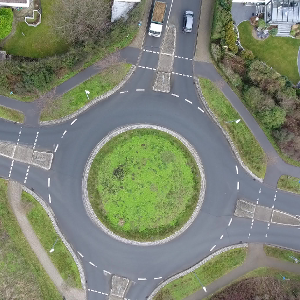}
       \caption*{$\text{rdb}_5$}
	\end{subfigure}
	\begin{subfigure}{0.32\linewidth}
	   \includegraphics[width=\linewidth]{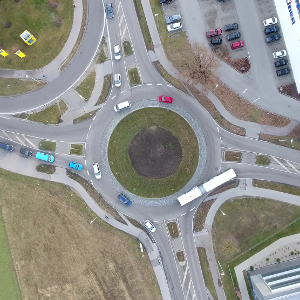}
       \caption*{$\text{rdb}_6$}
	\end{subfigure}
    {
    \begin{center}
        \begin{subfigure}{0.32\linewidth}
        \centering
        \includegraphics[width=\linewidth]{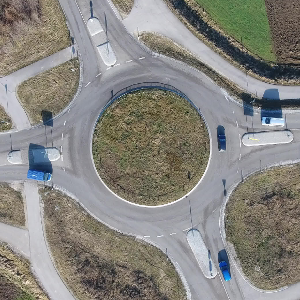}
        \caption*{$\text{rdb}_7$}
        \end{subfigure}
    \end{center}
    }
  \caption{
  An overview of the seven roundabouts included in the \texttt{openDD} dataset, with their respective abbreviation $\text{rdb}_i$ used in the dataset, $i \in \{1,2,\dots,7\}$.
    }
   \label{fig:datasetOverview}
\end{figure}

This work introduces the \texttt{openDD} dataset, a trajectory dataset recorded from a drone perspective.
The dataset includes $R=501$ recordings, each representing one coherent drone flight, capturing one of the $I = 7$ roundabouts covered in the dataset, depicted in \autoref{fig:datasetOverview}.
Each recording indexed by $r \in \mathcal R = \{1,\dots, R\}$ spans $5$ to $15$ minutes in total and was taken from the drone perspective with a camera capturing $30$~fps.
The used drone is a \texttt{DJI Phantom 4}, a high-end consumer drone, recording at a resolution of $3840\times2160$ pixels, being slightly below 4K.
The video footage taken by the drone is stabilized and rectified before it is used to detect and track all traffic participants in the given scene.

For each recording $r$ we define $N_r$ to be the number of time instants included in the recording, equal to the number of frames captured in the recorded video.

The \texttt{openDD} dataset defines which objects, each with unique object index $j$ are present at time instant \mbox{$n \in \{1, \dots, N_r\}$}.
The state vector $\mathbf{s}_{n}^{(j)}$ of an object $j$ at a time instant $n$ is defined by ($[]^T$ being the transpose)
\begin{equation}
  \begin{split}
    \mathbf{s}_{n}^{(j)} = [&\begin{matrix} x_{n}^{(j)}, & y_{n}^{(j)}, & \alpha_{n}^{(j)}, & w_{n}^{(j)}, & l_{n}^{(j)},\end{matrix} \\
    & \begin{matrix}v_{n}^{(j)}, & a_{\mathbf{l},n}^{(j)}, & a_{\mathbf{t},n}^{(j)}, & a_{n}^{(j)}, & c_{n}^{(j)}\end{matrix}]^T.
  \end{split}
\end{equation}
The vector $\left[x_{n}^{(j)}, y_{n}^{(j)}\right]^T$ describes the \gls{UTM} coordinates of the object's bounding box center.
The orientation of the bounding box is given by its yaw $\alpha_{n}^{(j)}$ in radiants relative to the x-axis of the \gls{UTM} reference coordinate system, whereas the dimension of the bounding box is given by the width $w_{n}^{(j)}$ and length $l_{n}^{(j)}$.
The dynamic state of the object is described by the velocity $v_n^{(j)}$ defined in \si{\meter\per\second}, whereas $a_{\mathbf{l},n}^{(j)}, a_{\mathbf{t},n}^{(j)}$, and $a_{n}^{(j)}$ describe the \textbf{l}ateral, \textbf{t}angential, and total acceleration of the object in \si{\meter\per\square\second}.
Additionally, the class $c_n^{(j)} \in \mathcal C = \{C, V, T, B, P, R, M, Y\}$ of the object is defined, with each object either being a passenger \textbf{c}ar, \textbf{v}an, \textbf{t}ruck, \textbf{b}us, \textbf{p}edestrian, t\textbf{r}ailer, \textbf{m}otorcycle, or bic\textbf{y}clist.
A visualization of the included bounding box information and the color-encoded class labels for a given scene can be seen in \autoref{fig:visualizedTrajectories}.

%Since trailers are towed by another vehicle, their object state (vector) does not include dynamic information and defines only the bounding box information, class $c_n^{(j)}$, and the object index $t_n^{(j)}$ of the object that tows the trailer, according to
%\begin{equation}
%  \begin{split}
%    \mathbf{s}_{n}^{(j)} = [&\begin{matrix} x_{n}^{(j)}, & y_{n}^{(j)}, & \alpha_{n}^{(j)}, & w_{n}^{(j)}, &l_{n}^{(j)}, & t_{n}^{(j)}, &c_{n}^{(j)}\end{matrix}]^T \\
%  \end{split}
%\end{equation}
% The speed and acceleration values of a given trailer $j$ can be approximated by the speed and acceleration values of the tow vehicle given in the tow vehicle's object state $\mathbf{s}_{n}^{\left(t_{n}^{(j)}\right)}$.

\begin{figure}[t] \includegraphics[width=\linewidth]{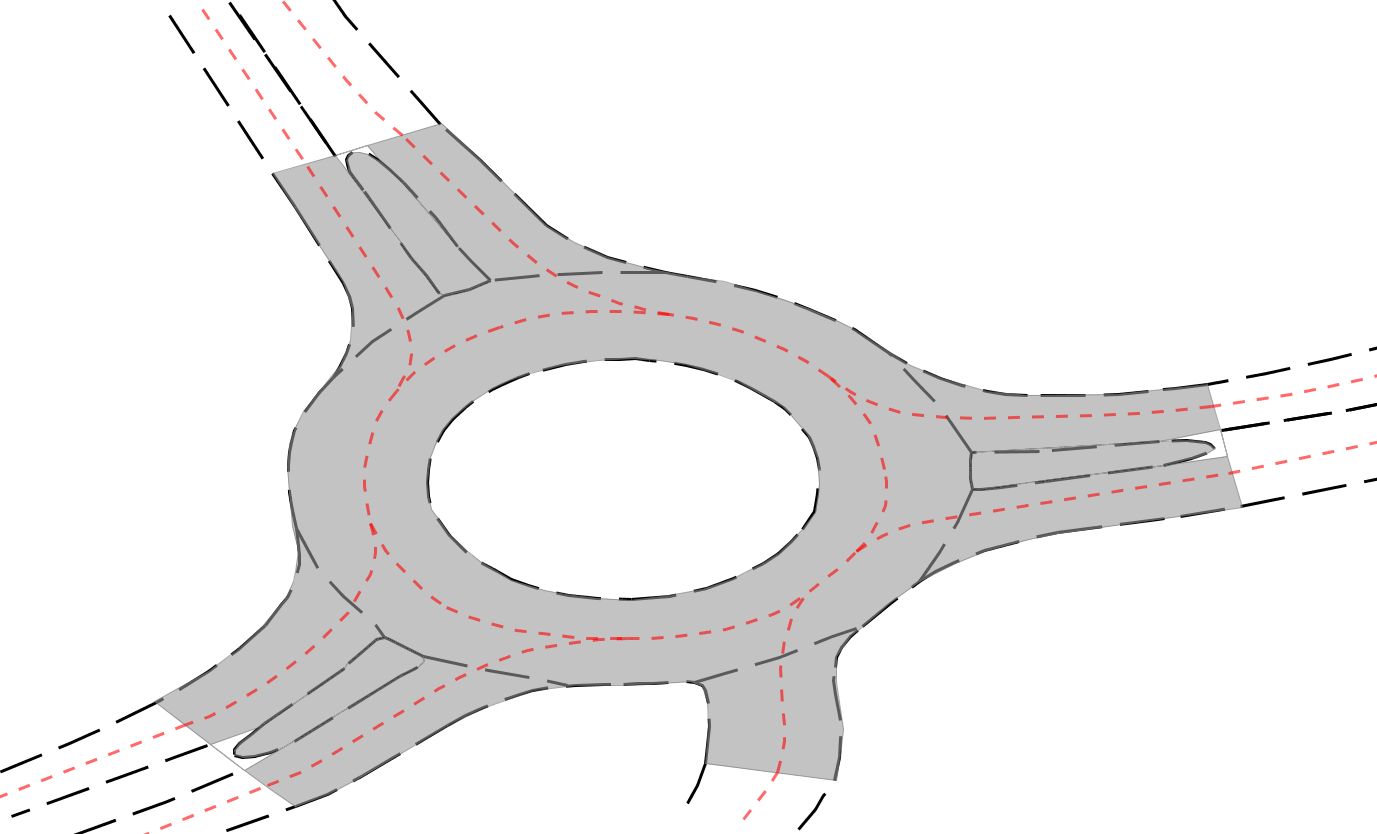} \caption{
    An exemplary visualisation of the shapefile for $\text{rdb}_1$ included in the dataset, created by the open source \gls{GIS} \texttt{OpenJUMP}.
    The centerlines are decoded in red dashed lines, whereas the lane boundaries are drawn in black dashed lines.
    The grey surface visualizes the polygon marked as \textit{drivable area}.
  }
  \label{fig:mapExample}
\end{figure}

We provide the underlying HD map for each roundabout $\text{rdb}_i$ with $i \in\{1,\dots,I\}$ included in the dataset.
The map data is provided as shapefiles and an \gls{XML} file and distinguishes three logical elements of the road topology: lane centerlines, lane boundaries, and drivable areas.
For each such logical element, three shapefiles are provided, a \texttt{.shp}, a \texttt{.shx}, and a \texttt{.dbf}, resulting in a total of nine shapefiles included in the dataset per roundabout.
The \texttt{.shp} file defines the underlying geometry of the logical element, such as the points of the lanes. 
The \texttt{.dbf} attribute files are in \texttt{dBase} format and define element-specific attributes.
For the lane centerlines, the \texttt{.dbf} file contains, among other attributes, a unique identifier for each lane centerline and a list of succeeding lane centerlines, preceding lane centerlines, and parallel lane centerlines.
The \texttt{.dbf} file for the lane boundaries defines the corresponding lane identifier, as well as the material of the lane boundary, such as \texttt{CONCRETE} for a curbstone and \texttt{NONE} for an implicit lane boundary.
The \texttt{.shx} file is an index file of the shape geometry \texttt{.shp} file, providing a way to quickly iterate over the defined geometry.
An exemplary visualisation of the provided shapefiles is given in \autoref{fig:mapExample}.
In addition to the shapefiles we also provide an \texttt{.\gls{XML}} file for each intersection $i$, representing the information contained in the shapefiles in a non-binary format that can be easily parsed by most programming languages.

Beyond the HD map and object states, the dataset includes a geo-referenced and anonymized example picture taken from the drone perspective of each intersection $i$.
This picture both provides a way of visualizing the trajectory data, as well as a supplementary input to the provided HD map.

\subsection{Dataset Statistics} \label{sec:datasetStatistics}
The dataset was recorded at different times of day, for each roundabout including at least \SI{2}{\hour} at the rush hour times in the morning and afternoon, as well as regular intervals in between rush hours.

The dataset spans \SI{62,7}{\hour}, of which \SI{18,8}{\hour} cover the first roundabout $\text{rdb}_1$ and the remaining \SI{43,9}{\hour} are distributed among the remaining six roundabouts $\text{rdb}_2$ to $\text{rdb}_7$, with varying lengths around \SI{7}{\hour} for each.
An example picture of each roundabout is depicted in \autoref{fig:datasetOverview}.
In total \SI{84774}{} trajectories are included in the dataset, covering \SI{8501,14}{\kilo\meter}.
A detailed overview of the \texttt{openDD} dataset, distinguishing between the seven different roundabouts included in the dataset, is provided in \autoref{tab:datasetOverview}.
Here, the number of trajectories, the average trajectory duration, length, velocity, and total acceleration is stated for each object class, as well as for each roundabout.
The average trajectory duration over all classes and all data subsets is \SI{17,64}{\second}, with an average trajectory length of \SI{100,28}{\meter}.

%As mentioned above, no dynamic information is given for trailers, since they don't move independently, but are towed by another object, which is referenced in their object state.
%Thus, trailers are not considered in the computation of the average velocity of \SI{6,00}{\meter\per\second} and average total acceleration of \SI{1,34}{\meter\per\square\second}.
The average velocity is \SI{6,63}{\meter\per\second} and the average total acceleration is \SI{1,42}{\meter\per\square\second}.

The relatively high amount of vehicles, \SI{81372} across the whole dataset, compared to the \SI{3402} pedestrians and bicyclists, is caused by the high percentage of covered rush hour times, as well as the remote locations of some of the covered roundabouts.
The roundabout $\text{rdb}_6$ has an especially high traffic load with \SI{13644} unique vehicles passing the roundabout in \SI{6,9}{\hour}.

Roundabout $\text{rdb}_2$ has a very high average trajectory duration of pedestrians, with \SI{65,42}{\second} compared to the average pedestrian trajectory duration of \SI{87,64}{\second}.
%At the same time the average trajectory length of \SI{66,11}{\meter} lies below the average of \SI{80,44}{\meter}.
The high pedestrian trajectory duration of $\text{rdb}_2$ is caused by several pedestrians idling in the recordings of $\text{rdb}_2$.

\section{Using the Dataset} \label{sec:Challenges}
Publications of trajectory prediction models that use this dataset should be evaluated in a uniform fashion.
To this end we define metrics to evaluate predicted trajectories, different splits of the \texttt{openDD} dataset, and propose several challenges using this dataset in the following.

\begin{table*}[t]
  \caption{Statistics of the \texttt{openDD} dataset, distinguishing the seven included roundabouts.} \label{tab:datasetOverview}
  \begin{center}
\begin{tabular}{r|r|r|r|r|r|r|r||r}
    Data Subset & $\text{rdb}_1$ & $\text{rdb}_2$ & $\text{rdb}_3$ & $\text{rdb}_4$ & $\text{rdb}_5$ & $\text{rdb}_6$ & $\text{rdb}_7$ & all \\
    \hline
    \hline
    Recorded time & \SI{18,8}{\hour} & \SI{7,5}{\hour} & \SI{7,0}{\hour} & \SI{7,7}{\hour} & \SI{7,1}{\hour} & \SI{6,9}{\hour} & \SI{7,7}{\hour} & \SI{62,7}{\hour} \\
    \hline
    \hline
    \# drone flights & \SI{153}{} & \SI{56}{} & \SI{54}{} & \SI{69}{} & \SI{60}{} & \SI{52}{} & \SI{57}{} & \SI{501}{} \\
    \hline
    \hline
    \multicolumn{1}{l|}{\# trajectories} & & & & & & & & \\
    passenger \textbf{c}ars & \SI{26879}{} & \SI{7685}{} & \SI{7100}{} & \SI{5983}{} & \SI{3510}{} & \SI{11730}{} & \SI{6512}{} & \SI{69399}{} \\
                    \textbf{v}an & \SI{2630}{} & \SI{740}{} & \SI{676}{} & \SI{497}{} & \SI{396}{} & \SI{923}{} & \SI{782}{} & \SI{6644}{} \\
                    \textbf{t}ruck & \SI{347}{} & \SI{420}{} & \SI{311}{} & \SI{79}{} & \SI{88}{} & \SI{484}{} & \SI{394}{} & \SI{2123}{} \\
                        \textbf{b}us & \SI{551}{} & \SI{79}{} & \SI{61}{} & \SI{76}{} & \SI{38}{} & \SI{78}{} & \SI{49}{} & \SI{932}{} \\
                \textbf{p}edestrian & \SI{963}{} & \SI{16}{} & \SI{38}{} & \SI{50}{} & \SI{122}{} & \SI{607}{} & \SI{23}{} & \SI{1819}{} \\
                        t\textbf{r}ailer & \SI{529}{} & \SI{367}{} & \SI{240}{} & \SI{71}{} & \SI{81}{} & \SI{393}{} & \SI{332}{} & \SI{2013}{} \\
                        \textbf{m}otorcycle & \SI{143}{} & \SI{11}{} & \SI{14}{} & \SI{32}{} & \SI{9}{} &  \SI{36}{} & \SI{16}{} & \SI{261}{} \\
            bi\textbf{y}clist & \SI{831}{} & \SI{52}{} & \SI{119}{} & \SI{96}{} & \SI{123}{} & \SI{332}{} & \SI{30}{} & \SI{1583}{} \\
                \hline
                            & \SI{32873}{} & \SI{9370}{} & \SI{8559}{} & \SI{6884}{} & \SI{4367}{} & \SI{14583}{} & \SI{8138}{} & \SI{84774}{} \\
                \hline \hline
    \multicolumn{1}{l|}{Average trajectory duration} & & & & & & & & \\
    passenger \textbf{c}ars & \SI{18,05}{\second} & \SI{16,01}{\second} & \SI{12,69}{\second} & \SI{15,74}{\second} & \SI{11,97}{\second} & \SI{18,54}{\second} & \SI{13,96}{\second} & \SI{16,47}{\second} \\
                \textbf{v}an & \SI{17,25}{\second} & \SI{16,43}{\second} & \SI{12,38}{\second} & \SI{16,16}{\second} & \SI{12,52}{\second} & \SI{18,62}{\second} & \SI{14,45}{\second} & \SI{16,16}{\second} \\
            \textbf{t}ruck & \SI{17,67}{\second} & \SI{17,97}{\second} & \SI{11,58}{\second} & \SI{18,40}{\second} & \SI{14,75}{\second} & \SI{16,80}{\second} & \SI{17,21}{\second} & \SI{16,46}{\second} \\
                \textbf{b}us & \SI{17,44}{\second} & \SI{17,44}{\second} & \SI{16,76}{\second} & \SI{21,45}{\second} & \SI{12,70}{\second} & \SI{16,58}{\second} & \SI{14,75}{\second} & \SI{17,32}{\second} \\
        \textbf{p}edestrian & \SI{63,69}{\second} & \SI{82,17}{\second} & \SI{76,76}{\second} & \SI{73,13}{\second} & \SI{50,66}{\second} & \SI{70,24}{\second} & \SI{41,80}{\second} & \SI{65,42}{\second} \\
            t\textbf{r}ailer & \SI{17,79}{\second} & \SI{18,27}{\second} & \SI{11,86}{\second} & \SI{17,82}{\second} & \SI{13,91}{\second} & \SI{17,38}{\second} & \SI{16,90}{\second} & \SI{16,79}{\second} \\
        \textbf{m}otorcycle & \SI{16,89}{\second} & \SI{14,35}{\second} & \SI{17,51}{\second} & \SI{15,35}{\second} & \SI{12,10}{\second} & \SI{16,76}{\second} & \SI{12,51}{\second} & \SI{16,18}{\second} \\
        bi\textbf{y}clist & \SI{23,13}{\second} & \SI{19,84}{\second} & \SI{24,70}{\second} & \SI{22,07}{\second} & \SI{21,50}{\second} & \SI{26,68}{\second} & \SI{18,00}{\second} & \SI{23,60}{\second} \\
    \hline
                            & \SI{19,43}{\second} & \SI{16,37}{\second} & \SI{13,09}{\second} & \SI{16,39}{\second} & \SI{13,47}{\second} & \SI{20,78}{\second} & \SI{14,38}{\second} & \SI{17,64}{\second} \\
    \hline
    \hline

    \multicolumn{1}{l|}{Average trajectory length} & & & & & & & & \\
    passenger \textbf{c}ars & \SI{96,17}{\meter} & \SI{120,77}{\meter} & \SI{80,76}{\meter} & \SI{111,55}{\meter} & \SI{92,83}{\meter} & \SI{108,81}{\meter} & \SI{108,27}{\meter} & \SI{101,75}{\meter} \\
                \textbf{v}an & \SI{93,29}{\meter} & \SI{120,21}{\meter} & \SI{77,60}{\meter} & \SI{109,65}{\meter} & \SI{93,34}{\meter} & \SI{106,53}{\meter} & \SI{107,69}{\meter} & \SI{99,45}{\meter} \\
            \textbf{t}ruck & \SI{86,66}{\meter} & \SI{104,42}{\meter} & \SI{59,28}{\meter} & \SI{108,42}{\meter} & \SI{91,54}{\meter} & \SI{87,24}{\meter} & \SI{105,98}{\meter} & \SI{90,89}{\meter} \\
                \textbf{b}us & \SI{84,20}{\meter} & \SI{106,05}{\meter} & \SI{78,16}{\meter} & \SI{106,27}{\meter} & \SI{67,09}{\meter} & \SI{81,70}{\meter} & \SI{83,46}{\meter} & \SI{86,51}{\meter} \\
        \textbf{p}edestrian & \SI{71,24}{\meter} & \SI{110,24}{\meter} & \SI{96,12}{\meter} & \SI{90,33}{\meter} & \SI{65,77}{\meter} & \SI{101,00}{\meter} & \SI{75,65}{\meter} & \SI{82,25}{\meter} \\
            t\textbf{r}ailer & \SI{84,34}{\meter} & \SI{103,96}{\meter} & \SI{57,86}{\meter} & \SI{99,54}{\meter} & \SI{82,74}{\meter} & \SI{86,33}{\meter} & \SI{103,38}{\meter} & \SI{88,76}{\meter} \\
        \textbf{m}otorcycle & \SI{93,09}{\meter} & \SI{128,27}{\meter} & \SI{90,62}{\meter} & \SI{105,43}{\meter} & \SI{93,83}{\meter} & \SI{96,49}{\meter} & \SI{107,03}{\meter} & \SI{97,30}{\meter} \\
        bi\textbf{y}clist & \SI{97,30}{\meter} & \SI{97,75}{\meter} & \SI{97,39}{\meter} & \SI{94,95}{\meter} & \SI{81,13}{\meter} & \SI{99,43}{\meter} & \SI{81,14}{\meter} & \SI{96,06}{\meter} \\
    \hline
                    & \SI{94,74}{\meter} & \SI{119,08}{\meter} & \SI{79,38}{\meter} & \SI{110,78}{\meter} & \SI{91,36}{\meter} & \SI{106,63}{\meter} & \SI{107,56}{\meter} & \SI{100,28}{\meter} \\
                    \hline
                    \hline
    \multicolumn{1}{l|}{Average velocity} & & & & & & & & \\
    passenger \textbf{c}ars & \SI{6,00}{\meter\per\second} & \SI{8,13}{\meter\per\second} & \SI{6,88}{\meter\per\second} & \SI{7,37}{\meter\per\second} & \SI{8,03}{\meter\per\second} & \SI{6,66}{\meter\per\second} & \SI{8,25}{\meter\per\second} & \SI{6,87}{\meter\per\second} \\
                \textbf{v}an & \SI{6,03}{\meter\per\second} & \SI{7,83}{\meter\per\second} & \SI{6,69}{\meter\per\second} & \SI{6,97}{\meter\per\second} & \SI{7,69}{\meter\per\second} & \SI{6,27}{\meter\per\second} & \SI{7,83}{\meter\per\second} & \SI{6,71}{\meter\per\second} \\
            \textbf{t}ruck & \SI{5,39}{\meter\per\second} & \SI{6,36}{\meter\per\second} & \SI{5,33}{\meter\per\second} & \SI{6,09}{\meter\per\second} & \SI{6,41}{\meter\per\second} & \SI{6,25}{\meter\per\second} & \SI{6,47}{\meter\per\second} & \SI{6,04}{\meter\per\second} \\
                \textbf{b}us & \SI{5,28}{\meter\per\second} & \SI{6,48}{\meter\per\second} & \SI{5,06}{\meter\per\second} & \SI{5,34}{\meter\per\second} & \SI{5,62}{\meter\per\second} & \SI{5,26}{\meter\per\second} & \SI{5,79}{\meter\per\second} & \SI{5,41}{\meter\per\second} \\
        \textbf{p}edestrian & \SI{1,27}{\meter\per\second} & \SI{1,85}{\meter\per\second} & \SI{1,38}{\meter\per\second} & \SI{1,56}{\meter\per\second} & \SI{1,39}{\meter\per\second} & \SI{1,47}{\meter\per\second} & \SI{1,96}{\meter\per\second} & \SI{1,37}{\meter\per\second} \\
        t\textbf{r}ailer & \SI{5,32}{\meter\per\second} & \SI{6,22}{\meter\per\second} & \SI{5,09}{\meter\per\second} & \SI{5,94}{\meter\per\second} & \SI{6,30}{\meter\per\second} & \SI{6,05}{\meter\per\second} & \SI{6,53}{\meter\per\second} & \SI{5,86}{\meter\per\second} \\
        \textbf{m}otorcycle & \SI{5,93}{\meter\per\second} & \SI{9,19}{\meter\per\second} & \SI{6,26}{\meter\per\second} & \SI{7,09}{\meter\per\second} & \SI{7,90}{\meter\per\second} & \SI{6,57}{\meter\per\second} & \SI{9,11}{\meter\per\second} & \SI{6,58}{\meter\per\second} \\
        bi\textbf{y}clist & \SI{4,64}{\meter\per\second} & \SI{5,32}{\meter\per\second} & \SI{4,30}{\meter\per\second} & \SI{4,70}{\meter\per\second} & \SI{4,15}{\meter\per\second} & \SI{4,09}{\meter\per\second} & \SI{5,02}{\meter\per\second} & \SI{4,50}{\meter\per\second} \\
    \hline
                    & \SI{5,80}{\meter\per\second} & \SI{7,91}{\meter\per\second} & \SI{6,69}{\meter\per\second} & \SI{7,21}{\meter\per\second} & \SI{7,62}{\meter\per\second} & \SI{6,33}{\meter\per\second} & \SI{8,01}{\meter\per\second} & \SI{6,63}{\meter\per\second} \\
                    \hline
                    \hline
    \multicolumn{1}{l|}{Average acceleration} & & & & & & & & \\
    passenger \textbf{c}ars & \SI{1,16}{\meter\per\square\second} & \SI{1,73}{\meter\per\square\second} & \SI{1,57}{\meter\per\square\second} & \SI{1,75}{\meter\per\square\second} & \SI{1,78}{\meter\per\square\second} & \SI{1,65}{\meter\per\square\second} & \SI{1,82}{\meter\per\square\second} & \SI{1,49}{\meter\per\square\second} \\
                \textbf{v}an & \SI{1,09}{\meter\per\square\second} & \SI{1,60}{\meter\per\square\second} & \SI{1,42}{\meter\per\square\second} & \SI{1,63}{\meter\per\square\second} & \SI{1,61}{\meter\per\square\second} & \SI{1,46}{\meter\per\square\second} & \SI{1,61}{\meter\per\square\second} & \SI{1,37}{\meter\per\square\second} \\
            \textbf{t}ruck & \SI{1,01}{\meter\per\square\second} & \SI{1,26}{\meter\per\square\second} & \SI{1,18}{\meter\per\square\second} & \SI{1,29}{\meter\per\square\second} & \SI{1,28}{\meter\per\square\second} & \SI{1,37}{\meter\per\square\second} & \SI{1,24}{\meter\per\square\second} & \SI{1,23}{\meter\per\square\second} \\
            \textbf{b}us & \SI{0,87}{\meter\per\square\second} & \SI{1,27}{\meter\per\square\second} & \SI{0,98}{\meter\per\square\second} & \SI{0,95}{\meter\per\square\second} & \SI{1,07}{\meter\per\square\second} & \SI{1,16}{\meter\per\square\second} & \SI{1,22}{\meter\per\square\second} & \SI{0,97}{\meter\per\square\second} \\
        \textbf{p}edestrian & \SI{0,18}{\meter\per\square\second} & \SI{0,30}{\meter\per\square\second} & \SI{0,19}{\meter\per\square\second} & \SI{0,24}{\meter\per\square\second} & \SI{0,22}{\meter\per\square\second} & \SI{0,21}{\meter\per\square\second} & \SI{0,24}{\meter\per\square\second} & \SI{0,20}{\meter\per\square\second} \\
        t\textbf{r}ailer & \SI{0,89}{\meter\per\square\second} & \SI{1,25}{\meter\per\square\second} & \SI{1,21}{\meter\per\square\second} & \SI{1,23}{\meter\per\square\second} & \SI{1,25}{\meter\per\square\second} & \SI{1,39}{\meter\per\square\second} & \SI{1,30}{\meter\per\square\second} & \SI{1,18}{\meter\per\square\second} \\
        \textbf{m}otorcycle & \SI{1,01}{\meter\per\square\second} & \SI{1,50}{\meter\per\square\second} & \SI{1,18}{\meter\per\square\second} & \SI{1,45}{\meter\per\square\second} & \SI{1,45}{\meter\per\square\second} & \SI{1,55}{\meter\per\square\second} & \SI{1,91}{\meter\per\square\second} & \SI{1,24}{\meter\per\square\second} \\
        bi\textbf{y}clist & \SI{0,60}{\meter\per\square\second} & \SI{0,63}{\meter\per\square\second} & \SI{0,77}{\meter\per\square\second} & \SI{0,81}{\meter\per\square\second} & \SI{0,78}{\meter\per\square\second} & \SI{0,87}{\meter\per\square\second} & \SI{0,71}{\meter\per\square\second} & \SI{0,70}{\meter\per\square\second} \\
    \hline
                    & \SI{1,10}{\meter\per\square\second} & \SI{1,67}{\meter\per\square\second} & \SI{1,51}{\meter\per\square\second} & \SI{1,69}{\meter\per\square\second} & \SI{1,67}{\meter\per\square\second} & \SI{1,54}{\meter\per\square\second} & \SI{1,74}{\meter\per\square\second} & \SI{1,42}{\meter\per\square\second} \\
\end{tabular}
  \end{center}
\end{table*}

\subsection{Distance Metrics}
Similar to our previous work~\cite{breuer_analysis_2019}, we define several distance metrics $D\big(\mathcal T^{(j)}, \bar{\mathcal T}^{(j)}\big)$ that can be used to evaluate the accuracy of a trajectory prediction algorithm.
For a given object with index $j$, this distance metric $D$ compares the predicted trajectory $\mathcal T^{(j)}$ with the actual ground truth trajectory $\bar{\mathcal T}^{(j)}$ of the object, as given in the dataset.
In the example scripts that are made available with the dataset, implementations of the used metrics are provided.

\textbf{Euclidean displacement at time $t_n$}:
The Euclidean point-to-point distance between the {$n$-th} trajectory point of $\mathcal T^{(j)}$ and the $m$-th trajectory point of $\bar{\mathcal T}^{(j)}$ is defined as
\begin{equation} \label{eq:euclideanDisplacement}
   \begin{split}
      &D_{{\mathrm{Et}}}\bigg(\mathcal T^{(j)}(n), \bar{\mathcal T}^{(j)}(m)\bigg) \\
      &~~~~~~~= \sqrt{{\Big(x_n^{(j)}-\bar{x}_m^{(j)}\Big)^2
      +{{\Big(y_n^{(j)}-\bar{y}_m^{(j)}\Big)}^2}}}.
   \end{split}
\end{equation}

\textbf{Mean-squared Euclidean distance}: Given two trajectories $\mathcal T^{(j)}, \bar{\mathcal T}^{(j)}$, spanning the same sequence of time instants $n \in \mathcal N = \{0,1,\dots,N\!\!-\!\!1\}$, the mean squared Euclidean distances \textit{between the two entire trajectories} is defined as the normalized sum of the squared Euclidean distances between the points corresponding to the same time instant $n$:
\begin{equation} \label{eq:meanSquaredEuclideanDisplacement}
   \begin{split}
     D_{{\mathrm{MSE}}}&\Big(\mathcal T^{(j)},\bar{\mathcal T}^{(j)}\Big) \\
      &= \frac{1}{N} \sum\limits_{n \in \mathcal N}D_{{\mathrm{Et}}}^2\Big(\mathcal T^{(j)}(n), \bar{\mathcal T}^{(j)}(n)\Big). \\
      % = \frac{1}{N} \sum\limits_{n \in \mathcal N}&{\Big(x_n^{(j)}-x_n^{(j)}\Big)^2 +{{\Big(y_n^{(j)}-y_n^{(j)}\Big)}^2}}.
   \end{split}
\end{equation}

\textbf{Modified Hausdorff (MH) distance}: The following definition of the \gls{MH} distance is adopted from a work on object matching by Dubuisson \textit{et al.}~\cite{dubuisson_modified_1994}.

The definition of the \textit{point-to-set distance} between the \mbox{$n$-th} point of a trajectory $\mathcal T^{(j)}(n)$, and another entire trajectory $\bar{\mathcal T}^{(j)}$, is:
\begin{equation}
D_{\mathrm{PS}}\Big(\mathcal T^{(j)}(n), \bar{\mathcal T}^{(j)}\Big) = \min_{m \in \mathcal N} \bigg(D_{\mathrm{Et}}\Big(\mathcal T^{(j)}(n),\bar{\mathcal T}^{(j)}(m)\Big)\bigg).
\end{equation}
\begin{table*}[!ht]
  \caption{Description of the three training data splits $\mathcal R_1, \mathcal R_2, \mathcal R_3$, as well as of the test data splits $\mathcal R_A, \mathcal R_B, \mathcal R_C$, as defined for the challenges described in \autoref{sec:Challenges}.} \label{tab:datasetSplits}
  \begin{center}
  \begin{tabular}{r||r||r|r||r|r||r|r||r||r||r||r}
    Subset & \multicolumn{1}{c||}{$\mathcal R_{123}$} & \multicolumn{2}{c||}{$\mathcal R_1$} & \multicolumn{2}{c||}{$\mathcal R_2$} & \multicolumn{2}{c||}{$\mathcal R_3$} & $\mathcal R_{ABC}$ & \multicolumn{3}{c}{}  \\
& & $\text{train}$ & $\text{val}$ & $\text{train}$ & $\text{val}$ & $\text{train}$ & $\text{val}$ & & $\mathcal R_A$ & $\mathcal R_B$ & $\mathcal R_C$ \\
\hline
\hline
Recorded time & \SI{47,3}{\hour} & \SI{38,8}{\hour} & \SI{8,5}{\hour} & \SI{42,1}{\hour} & \SI{5,7}{\hour} & \SI{34,0}{\hour} & \SI{7,5}{\hour} & \SI{15,4}{\hour} & \SI{8,5}{\hour} & \SI{7,0}{\hour} & \SI{9,4}{\hour} \\
\hline
\hline
\# recordings from & & & & & & & & & & \\
$\text{rdb}_1$ & 130 & 107 & 23 & 0 & 0 & 107 & 23 &  23 & 23 & 0 & 23 \\
$\text{rdb}_2$ & 48 & 40 & 8 & 40 &  8 & 40 & 8 & 8 & 8 & 0 & 8 \\
$\text{rdb}_3$ & 0 & 0 & 0 & 0 & 0 & 0 & 0 & 54 & 0 & 54 & 8 \\
$\text{rdb}_4$ & 59 & 49 & 10 & 49 & 10 & 49 & 10 & 10 & 10 & 0 & 10 \\
$\text{rdb}_5$ & 51 & 42 & 9 & 42 & 9 & 42 & 9 & 9 & 9 & 0 & 9 \\
$\text{rdb}_6$ & 44 & 36 & 8 & 36 & 8 & 0 & 0 & 8 & 8 & 0 & 8 \\
$\text{rdb}_7$ & 48 & 39 & 9 & 39 & 9 & 39 & 9 & 9 & 9 & 0 & 9  \\
\end{tabular}

  \end{center}
\end{table*}
\vspace*{-0.5cm}

The \textit{directed modified Hausdorff} (DMH) distance is defined by Dubuisson \textit{et al.}~\cite{dubuisson_modified_1994} as
\begin{equation}
  D_{\mathrm{DMH}}\Big(\mathcal T^{(j)}, \bar{\mathcal T}^{(j)}\Big) = \frac{1}{N}\sum\limits_{n \in \mathcal N} D_{\mathrm{PS}}\Big(\mathcal T^{(j)}(n), \bar{\mathcal T}^{(j)}\Big).
\end{equation}
The undirected, \textit{modified Hausdorff} (MH) distance is then computed by taking the maximum over the two directed distances:
\begin{equation} \label{eq:undirectedModifiedHausdorffDistance}
   \begin{split}
   D_{\mathrm{MH}}&\Big(\mathcal T^{(j)}, \bar{\mathcal T}^{(j)}\Big) = D_{\mathrm{MH}}\Big(\bar{\mathcal T}^{(j)}, \mathcal T^{(j)}\Big) \\
   = & \max\bigg(D_{\mathrm{DMH}}\Big(\mathcal T^{(j)}, \bar{\mathcal T}^{(j)}\Big), D_{\mathrm{DMH}}\Big(\bar{\mathcal T}^{(j)}, \mathcal T^{(j)}\Big)\bigg). \\
   \end{split}
\end{equation}
The \gls{MH} distance captures the spatial similarity of the trajectories without considering temporal misalignments.
For example, two trajectories encoding the same path traveled at different velocities would have a low \gls{MH} distance, while their Euclidean displacement $D_{\mathrm{Et}}$ and mean squared Euclidean distance $D_{\mathrm{MSE}}$ would be high.

% For example, two trajectories describing a right turn at an intersection with high discrepancy in velocity would result in a low \gls{MH} distance, while the Euclidean displacement $D_{\mathrm{Et}}$ and mean squared Euclidean distance $D_{\mathrm{MSE}}$ between these trajectory points would yield higher values.

\subsection{Dataset Splits} \label{sec:datasetSplit}
We divide the drone recordings of the seven roundabouts $\text{rdb}_i$ into subsets for training, validation, and testing, as specified in \autoref{tab:datasetSplits}.
The exact assignment of recordings to the different subsets introduced in the following, defining which recording belongs to which subset, is provided in the dataset.

For testing purposes, we use all recordings $r$ from $\text{rdb}_3$, as well as $15\%$ of the recordings from the other roundabouts making up the total test set $\mathcal R_{ABC} \subset \mathcal R$.
We define three different subsets of the total test set $\mathcal R_{ABC}$ to evaluate algorithms on: $\mathcal{R}_A$, $\mathcal{R}_B$, $\mathcal{R}_C \subset \mathcal{R}_{ABC}$.

The subset $\mathcal{R}_A$ includes all recordings in $\mathcal{R}_{ABC}$, but the ones of $\text{rdb}_3$, thus it includes $15\%$ from all roundabouts $\text{rdb}_i$ with $i \in \{1,2,4,5,6,7\}$.
The second subset $\mathcal{R}_B$ covers all recordings of $\text{rdb}_3$.
Lastly, $\mathcal{R}_C$ includes $15\%$ from the recordings of $\text{rdb}_3$, as well as all recordings from the other roundabouts included in $\mathcal{R}_{ABC}$, such that in total $15\%$ of each roundabout are covered by this test set.

For training purposes, from the entire data \mbox{$\mathcal R = \mathcal R_{ABC} \cup \mathcal R_{123}$} we split the recordings $\mathcal R_{123}$ not included in $\mathcal{R}_{ABC}$ into three different splits \mbox{$\mathcal R_k={\mathcal R}_k^{\text{train}} \cup {\mathcal R}^\text{val}_k, k \in \{1,2,3\}$}.
To define the splits $\mathcal R_k$, we split the data included in $\mathcal R_{123}$ for the roundabouts $\text{rdb}_i, i \in \{1,2,4,5,6,7\}$, such that around $18\%$ of the recordings of each roundabout, except $\text{rdb}_3$, form the validation set and the remaining $82\%$ the training set.
% This division corresponds to roughly $70\%$ of the total available data for a given roundabout to belong to the training set, and $15\%$ to the validation set.
For the first split $\mathcal R_1$ we include the training and validation subsets formed in this way for all the roundabouts included in $\mathcal R_{123}$.
The second split $\mathcal R_2$, and third split $\mathcal R_3$ are equal to $\mathcal R_1$, but leave out all recordings from $\text{rdb}_1$ and $\text{rdb}_6$, respectively.

An important feature of this proposed division of the dataset is that the splits $\mathcal R_{k}$ can be combined with any of the test sets $\mathcal R_A, \mathcal R_B,$ and $\mathcal R_C$ to analyze different aspects of the learning process.
The test set $\mathcal R_A$ only includes recordings for topologies that are also included in $\mathcal R_{123}$, allowing for an evaluation of the model's ability to predict trajectories for previously seen roundabout topologies.
The test set $\mathcal R_B$ includes only recordings from an unseen roundabout, thus it is suitable to evaluate the generalization capability of the learned model.
Especially a combination with the split $\mathcal R_3$ is of interest, since $\mathcal R_3$ does not include the data of $\text{rdb}_6$, which is the only other roundabout with a similar topology including lanes to skip the center roundabout lane.
Lastly, the test set $\mathcal R_C$ is a mixture between the other two test sets, enabling an evaluation of both the learning capacity of the model, as well as its generalization ability.

\subsection{Challenges}
We encourage all publications using the \texttt{openDD} dataset to report their results using the Euclidean displacement as defined in \autoref{eq:euclideanDisplacement}, and the \gls{MH} as defined in \autoref{eq:undirectedModifiedHausdorffDistance} at the maximal prediction horizon they are reporting, and after \SI{1}{\second}, \SI{3}{\second}, and \SI{6}{\second}, if applicable.

Along the lines of our previous work~\cite{breuer_analysis_2019}, we propose to investigate the utility of the various information of the environment provided in the dataset.
Thus, we invite parties interested in using the dataset to adopt their algorithms for trajectory prediction in such a way that they can work with variable input data.

Out of the possible nine combinations between the three test sets $\mathcal R_A, \mathcal R_B$, and $\mathcal R_C$ and the training and validation splits $\mathcal R_K$, we encourage researchers using the dataset to report the results of the following four combinations, with the first two being the most relevant.

\textbf{Train on $\mathbf{\mathcal R_1}$, test on $\mathbf{\mathcal R_A}$:} All roundabout topologies tested upon have been seen during the training.
Thus the results on this combination will measure the general capability of the model to solve the task at hand.

\textbf{Train on $\mathbf{\mathcal R_1}$, test on $\mathbf{\mathcal R_B}$:} The test set only includes data covering a previously unseen roundabout, this combination can be used to assess the generalization capability of the evaluated model.

\textbf{Train on $\mathbf{\mathcal R_3}$, test on $\mathbf{\mathcal R_B}$:} Only recordings from a previously unseen roundabout, $\text{rdb}_3$ are included in the test set.
Additionally, no recordings from the roundabout with the road topology most similar to the one of the test roundabout, $\text{rdb}_6$ is included in $\mathcal R_3$.
This combination represents an even more difficult generalization evaluation.

\textbf{Train on $\mathbf{\mathcal R_2}$, test on $\mathbf{\mathcal R_C}$:} No recordings from $\text{rdb}_1$, the roundabout which covers almost \SI{20}{\hour}, is included in $\mathcal R_2$.
Recordings from both seen and unseen roundabouts is included in the test set.
This allows for an evaluation of both the capability of the model as well as the generalization ability and has an almost balanced amount of hours covered for each roundabout in $\mathcal R_1$.

Beyond the reporting of the aforementioned measures and training/testing splits, we propose three further research challenges for trajectory prediction using the \texttt{openDD} dataset:
\begin{enumerate}
  \item Evaluation of the benefit of knowledge of the movement of other objects up to time instant $t$.
  \item Evaluation of the benefit of the provided map data, divided by centerlines, lane boundaries and drivable areas.
    Here, a separate evaluation with the same algorithm using all of the provided information, only centerlines, and only drivable areas is desirable.
  \item Given the image of each intersection $i \in \{1,\dots,I\}$ provided in the dataset is used by the trajectory prediction algorithm, we would like the authors to evaluate the benefit of the image.
    The image can be for example used as an input for a \gls{CNN}, and be combined with a birds-eye view of the current traffic scene, implicitly encoding the structure of the surroundings~\cite{breuer_analysis_2019}.
\end{enumerate}
If challenge 3) shows that the information provided by the image gives additional benefits for the trajectory prediction task, interesting follow-up research would deal with how to integrate this information into the HD map.

\section{FUTURE WORK} \label{sec:futureWork}
The \texttt{openDD} dataset introduced in this work is the biggest published trajectory dataset recorded from a drone perspective as of today.
In addition to its length of over \SI{62}{\hour}, it covers varying roundabout topologies, which makes it also valuable to study the generalization of trajectory prediction algorithms trained on it.
The license of the provided \texttt{openDD} dataset, covering commercial and non-commercial usage, makes it appealing to both research institutions, as well as to companies.
In a future publication we plan to provide a baseline on the given dataset, addressing the challenges introduced in \autoref{sec:Challenges}.

\section{ACKNOWLEDGMENT}
The authors would like to thank all partners within the H2020 co-funded project L3Pilot --- grant agreement number 723051 --- for their cooperation in making this dataset available online.
Responsibility for the information and views set out in this publication lies entirely with the authors.

   \nocite{*}
% Generated by IEEEtran.bst, version: 1.14 (2015/08/26)

\end{document}